\documentclass[11pt,a4paper]{article}
\usepackage[hyperref]{emnlp2020}
\usepackage{times}
\usepackage{latexsym}

\usepackage{linguex}
\usepackage{graphicx}
\usepackage{amsmath}
\usepackage{amssymb}
\usepackage{booktabs}

\usepackage{microtype}

\aclfinalcopy 

\title{BERTs of a feather do not generalize together: Large variability in generalization across models with similar test set performance}

\author{R. Thomas McCoy,\textsuperscript{1} Junghyun Min,\textsuperscript{1} and Tal Linzen\textsuperscript{2} \\
  \textsuperscript{1}Department of Cognitive Science, Johns Hopkins University \\
  \textsuperscript{2}Department of Linguistics and Center for Data Science, New York University \\
  \texttt{tom.mccoy@jhu.edu}, \texttt{jmin10@jhu.edu}, \texttt{linzen@nyu.edu} \\}

\date{}

\begin{document}
\maketitle
\begin{abstract}
If the same neural network architecture is trained multiple times on the same dataset, will it make similar linguistic generalizations across runs? To study this question, we fine-tuned 100 instances of BERT on the Multi-genre Natural Language Inference (MNLI) dataset and evaluated them on the HANS dataset, which evaluates syntactic generalization in natural language inference. 
  On the MNLI development set, 
  the behavior of all instances 
  was remarkably consistent, with accuracy ranging between 83.6\% and 84.8\%. In stark contrast, the same models varied widely in their generalization performance. For example, on the simple case of subject-object swap (e.g., determining that \textit{the doctor visited the lawyer} does not entail \textit{the lawyer visited the doctor}), accuracy ranged from 0.0\% to 66.2\%. 
  Such variation is likely due to the presence of many local minima in the loss surface that are equally attractive to a low-bias learner such as a neural network; decreasing the variability may therefore require 
  models with stronger inductive biases.
\end{abstract}

\setlength{\Exlabelwidth}{0.8em}
\setlength{\SubExleftmargin}{1.25em}

\section{Introduction}

Generalization is a crucial component of learning a language. No training set can contain all possible sentences, so learners must be able to generalize to sentences that they have never encountered before. We differentiate two types of generalization:

\begin{enumerate}
    \item \textbf{In-distribution generalization:} Generalization to examples which are novel but which are drawn from the same distribution as the training set.
    \item \textbf{Out-of-distribution generalization:} Generalization to examples drawn from a different distribution than the training set.
\end{enumerate}

\noindent
Standard test sets in natural language processing are
generated in the same way as the corresponding training set, therefore testing only in-distribution generalization. Current neural architectures perform very well at this type of generalization. For example, on the 
natural language understanding tasks included in the 
GLUE benchmark \cite{wang2019glue}, several Transformer-based models \cite{liu2019roberta,liu2019mtdnn,raffel2019exploring} have surpassed the human baselines from \newcite{nangia2019human}. 

However, this strong performance does not necessarily indicate mastery of language. Because of biases in training distributions, it is often possible for a model to achieve strong in-distribution generalization by using shallow heuristics rather than deeper linguistic knowledge. Therefore, evaluating only on standard test sets cannot reveal whether a model has learned abstract properties of language or if it has only learned shallow heuristics.

An alternative evaluation approach addresses this flaw by testing how the model handles particular linguistic phenomena, using datasets designed to be impossible to solve using shallow heuristics.
In this line of investigation, which tests out-of-distribution generalization, the results are more mixed. Some works have found successful handling of phenomena such as subject-verb agreement \citep{gulordava2018colorless} and filler-gap dependencies \citep{wilcox2018rnn}. Other works, however, have illuminated surprising failures even on seemingly simple types of examples \citep{marvinlinzen18,mccoy2019right}.
Such results make it clear that there is still much room for improvement in how neural models perform on syntactic structures that are rare in training corpora.

In this work, we investigate whether the linguistic generalization behavior of a given neural architecture is consistent across multiple instances of that architecture. This question is important because, in order to tell which types of architectures generalize best, we need to know whether successes and failures of generalization should be attributed to aspects of the architecture or to random luck in the choice of the model's initial weights.

We investigate this question using the task of natural language inference (NLI). We fine-tuned 100 instances of BERT \citep{devlin2018bert} on the MNLI dataset \citep{williams2018multinli}.\footnote{The weights for all 100 fine-tuned models are publicly available at \url{https://github.com/tommccoy1/hans}.} These 100 instances differed only in (i) the initial weights of the classifier trained on top of BERT, and (ii) the order in which training examples were presented. All other aspects of training, including the initial weights of BERT, were held constant. We evaluated these 100 instances on both the in-distribution MNLI development set and the out-of-distribution HANS evaluation set \citep{mccoy2019right}, which tests
syntactic generalization in NLI models.

We found that these 100 instances were remarkably consistent in their in-distribution generalization accuracy, with all accuracies on the MNLI development set falling in the range 83.6\% to 84.8\%, and with a high level of consistency on labels for specific examples (e.g., we identified 526 examples that all 100 instances labeled incorrectly). In contrast,
these 100 instances varied dramatically in their out-of-distribution generalization performance;
for example, on one of the thirty categories of examples in the HANS dataset, accuracy ranged from 4\% to 76\%. 
These results show that, when assessing the linguistic generalization of neural models, it is important to consider multiple training runs of each architecture, since models can differ vastly in how they perform on examples drawn from a different distribution than the training set, even when they perform similarly on an in-distribution test set.

\section{Background}

\subsection{In-distribution generalization}

Several works have noted that the same architecture can have very different in-distribution generalization across restarts of the same training process \citep{reimers2017reporting, reimers2018comparing, madhyastha2019model}.
Most relevantly for our work, 
fine-tuning of BERT is unstable for some datasets, such that some runs achieve state-of-the-art results while others perform poorly \cite{devlin2018bert,phang2018sentence}.
Unlike these past works, we focus on \textit{out-of-distribution} generalization, rather than in-distribution generalization.

\subsection{Out-of-distribution generalization}

Several other works have noted 
variation in out-of-distribution syntactic generalization.
\newcite{weber2018} trained 50 instances of a sequence-to-sequence model on a  
symbol replacement task.
These instances consistently
had above 99\% accuracy 
on the in-distribution test set but 
varied
on out-of-distribution generalization sets; 
in the most variable case, accuracy ranged from close to 0\% to over 90\%.
Similarly, \newcite{mccoy2018revisiting} trained 100 instances for each of six types of 
networks, using a synthetic training set that was ambiguous between two generalizations. 
Some models consistently made the same generalization across runs, but others varied considerably, with some instances of a given architecture strongly preferring one of the two  generalizations that were plausible given the training set, while other instances strongly preferred the other generalization. 
Finally, \newcite{livska2018memorize} trained 5000 instances of recurrent neural networks on the lookup tables task. Most of these 
instances failed on compositional generalization, but 
a small number 
generalized well. 

These works on variation in out-of-distribution generalization all used simple, synthetic tasks with training sets
designed to exclude certain types of examples.
Our work tests if models are still as variable when trained on a natural-language training set that is not adversarially designed. 
In concurrent work, \citet{zhou2020curse} also measured variability in out-of-distribution performance for 3 models (including BERT) on 12 datasets (including HANS). Their work has impressive breadth, whereas we instead aim for depth: We analyze the particular categories within HANS to give a fine-grained investigation of syntactic generalization, while \citeauthor{zhou2020curse} only report overall accuracy averaged across categories. In addition, we fine-tuned 100 instances of BERT, while \citeauthor{zhou2020curse} only fine-tuned 10 instances. The larger number of instances allows us to investigate the extent of the variability in more detail.

\subsection{Linguistic analysis of BERT}

Many recent papers have sought a deeper understanding of BERT, whether to assess its encoding of sentence structure \citep{lin2019open, hewitt2019structural, chrupala2019correlating, jawahar2019bert, tenney2018what}; its representational structure more generally \citep{abnar2019blackbox}; its handling of specific linguistic phenomena such as subject-verb agreement \citep{goldberg2019assessing}, negative polarity items \citep{warstadt2019investigating}, function words \citep{kim2019probing}, or a variety of psycholinguistic phenomena \citep{ettinger2019bert}; its internal workings \citep{coenen2019visualizing,tenney2019bert,clark2019does}; or its inductive biases \citep{warstadt2020can}. The novel contribution of this work is the focus on variability across a large number of fine-tuning runs; previous works have generally used models without fine-tuning or have used only a small number of fine-tuning runs (usually only one fine-tuning run, or at most ten fine-tuning runs).

\begin{figure*}
\resizebox{\textwidth}{!}{
\begin{tabular}{lp{6.5cm}p{6cm}} \toprule
    Heuristic & Definition & \multicolumn{1}{l}{Example}  \\ \midrule
    Lexical overlap & Assume that a premise entails all hypotheses constructed from words in the premise & \textbf{The doctor} was \textbf{paid} by \textbf{the actor}. \newline $\xrightarrow[\textsc{WRONG}]{}$ The doctor paid the actor. \\ 
    \midrule
    Subsequence & Assume that a premise entails all of its contiguous subsequences. & The doctor near \textbf{the actor danced}. \newline $\xrightarrow[\textsc{WRONG}]{}$ The actor danced. \\
     \midrule
    Constituent & Assume that a premise entails all complete subtrees in its parse tree. & If \textbf{the artist slept}, the actor ran. \newline  $\xrightarrow[\textsc{WRONG}]{}$ The artist slept.\\
    \bottomrule
\end{tabular}
}
\caption{The heuristics targeted by the HANS dataset, along with examples of incorrect entailment predictions that these heuristics would lead to. (Figure from \citealt{mccoy2019right}.)} \label{fig:hans}
\end{figure*}

\section{Method}

\subsection{Task and datasets}

We used the task of natural language inference (NLI, also known as Recognizing Textual Entailment; \citealp{condoravdi2003entailment,dagan2005pascal,dagan2013recognizing}), 
which involves giving a model two sentences, called the \textit{premise} and the \textit{hypothesis}. The model must then output 
\textit{entailment} if the premise entails (i.e., implies the truth of) the hypothesis, \textit{contradiction} if the premise contradicts the hypothesis, or \textit{neutral} otherwise. For training, we used the training set of the MNLI dataset \citep{williams2018multinli}, examples from which are given below: 

\ex. \a. \textbf{Premise:} Finally she turned back to him. 
\b. \textbf{Hypothesis:} She turned to him.
\c. \textbf{Label:} Entailment

\ex. \a. \textbf{Premise:} You outwitted me.
\b. \textbf{Hypothesis:} You have never outwitted me.
\c. \textbf{Label:} Contradiction

\ex. \a. \textbf{Premise:} okay well i live in Carrollton
\b. \textbf{Hypothesis:} I have a house in Carrollton. 
\c. \textbf{Label:} Neutral

To test in-distribution generalization, we used
the MNLI \texttt{matched} development set, which was generated in the same way as the MNLI training set. We used the development set 
rather than the test set 
because the test set labels are not available to the public. This development set was not used in any way 
during training, making it effectively a test set. To test out-of-distribution generalization, we used the HANS dataset \citep{mccoy2019right}, which contains NLI examples designed to require understanding of syntactic structure. More specifically, 
HANS targets three  structural heuristics that models trained on MNLI are likely to learn (for definitions and examples, see Figure~\ref{fig:hans}).

To assess whether a model has learned these heuristics, 
HANS contains examples where each heuristic makes the right predictions (i.e., where the correct label is \textit{entailment}) and examples where each heuristic makes the wrong predictions (i.e., where the correct label is \textit{non-entailment}). A model that has adopted one of the heuristics will output \textit{entailment} for all examples targeting that heuristic, even when the correct answer is \textit{non-entailment}. 

\subsection{Models and training}

All of our models consisted of BERT with a linear classifier on top of it outputting labels of \textit{entailment}, \textit{contradiction}, or \textit{neutral}.  We fine-tuned 100 instances of this model on MNLI using the fine-tuning code from the BERT GitHub repository.\footnote{\url{github.com/google-research/bert}} The BERT component of each instance was initialized with the pre-trained \texttt{bert-base-uncased} weights. For evaluation on HANS, 
we translated outputs of \textit{contradiction} and \textit{neutral} into  a single
\textit{non-entailment} label, following \newcite{mccoy2019right}.
The fine-tuning process
proceeded for 3 epochs 
and modified the weights of both the BERT component and the classifier. 
Following \newcite{devlin2018bert}, across 
fine-tuning 
runs we varied only (i) the random initial weights of the classifier and (ii) the order in which training examples were presented.
All other aspects, including the initial pre-trained weights of the BERT component, were held constant.

\section{Results}

\subsection{In-distribution generalization}

\begin{figure}
    \centering
    \includegraphics[width=0.48\columnwidth]{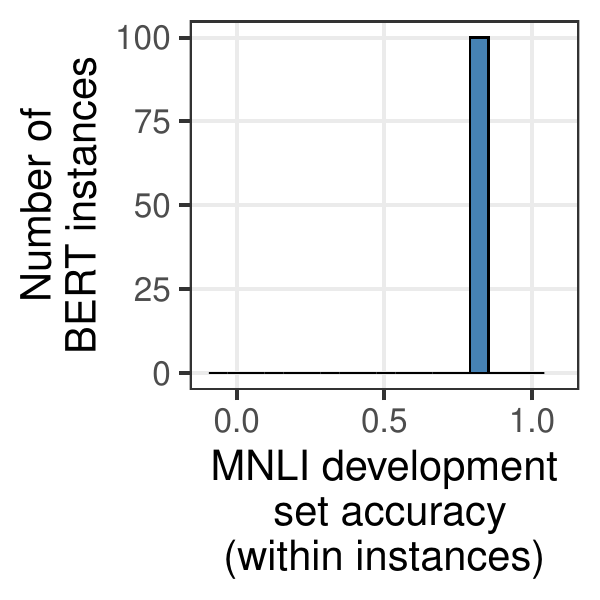} \hfill
    \includegraphics[width=0.48\columnwidth]{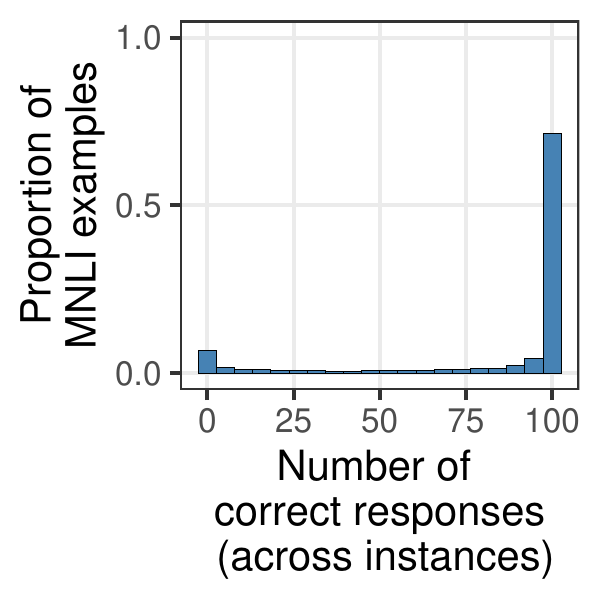}
    \caption{In-distribution generalization. Left: Within-instance accuracy on the MNLI development set; all BERT instances had scores near 84\%. Right: 
    Across-instance accuracy on individual examples in the MNLI development set; e.g., 66\% of the examples were answered correctly by all 100 instances.
    For numerical results, see Figure~\ref{tab:results_merged}.}
    \label{fig:mnli_results}
\end{figure}

The 100 instances were remarkably consistent
on in-distribution generalization, with all models 
scoring between 83.6\% and 84.8\% on the MNLI development set (Figure~\ref{fig:mnli_results}, left). 
Numerical statistics for the performance of our 100 instances of BERT on MNLI and HANS can be found in Figure~\ref{tab:results_merged}, and statistics for HANS broken down by linguistic construction can be found in Figures~\ref{tab:fine_grained_entailed} and~\ref{tab:fine_grained_nonentailed}. Finally, to see model-by-model results, see \url{https://github.com/tommccoy1/hans}. 

The instances were also highly consistent in their choice of labels for particular examples (Figure~\ref{fig:mnli_results}, right); in the rest of this subsection, we provide some quantitative and qualitative analysis of consistency of performance on individual examples. 
On average, among any pair of fine-tuned BERT instances, the two members of the pair agreed on the labels of 93.1\% of the examples (when considering all three labels of \textit{entailment}, \textit{contradiction}, and \textit{neutral}, rather than the collapsed labels of \textit{entailment} and \textit{non-entailment}). To give a sense of consistency across all 100 instances (rather than only among pairs of instances),
Figure~\ref{fig:mnli_results} (right) illustrates how consistent our 100 instances were on their answers to individual examples in the MNLI development set. Of the 9815 examples in the set, there were 6526 that all 100 instances labeled correctly, and 526 that all instances labeled incorrectly. Thus, the consistent score of about 84\% on the MNLI development set can be partially explained by the fact that there are certain examples that all models answered correctly or that all models answered incorrectly, as models were consistently correct or incorrect on 72\% of the examples.

Examples \ref{ex:easy1} through \ref{ex:easy3} show some of the 6526 cases that all 100 instances answered correctly:

\ex. \label{ex:easy1} \a. \textbf{Premise:} The new rights are nice enough
\b. \textbf{Hypothesis:} Everyone really likes the newest benefits
\c. \textbf{Label:} Neutral

\ex. \label{ex:easy2} \a. \textbf{Premise:} This site includes a list of all award winners and a searchable database of Government Executive articles.
\b. \textbf{Hypothesis:} The Government Executive articles housed on the website are not able to be searched.
\c. \textbf{Label:} Contradiction

\ex. \label{ex:easy3} \a. \textbf{Premise:} You and your friends are not welcome here, said Severn.
\b. \textbf{Hypothesis:} Severn said the people were not welcome there.
\c. \textbf{Label:} Entailment

\noindent
Examples \ref{ex:hard1} through \ref{ex:hard6} show some of the 526 cases that all 100 instances answered incorrectly. Some of these examples arguably have incorrect labels in the dataset, such as \ref{ex:hard1} (because the hypothesis mentions a report which the premise does not mention), so it is unsurprising that models found such examples difficult. Other consistently difficult examples involve areas that one might intuitively expect to be tricky for models trained on natural language, such as world knowledge (e.g., \ref{ex:hard2} requires knowledge of how long forearms are, and \ref{ex:hard3} requires knowledge of what nodding is), the ability to count (e.g., \ref{ex:hard4}), or fine-grained shades of meaning that might require multiple steps of reasoning (e.g., \ref{ex:hard5} and \ref{ex:hard6}). Some of the consistently difficult examples have a high degree of lexical overlap yet are not labeled \textit{entailment} (such as \ref{ex:hard7}); the difficulty of such examples adds further evidence to the conclusion that these models have adopted the lexical overlap heuristic. Finally, there are some examples, such as \ref{ex:hard8}, for which it is unclear why models find them so difficult.

\ex. \label{ex:hard1}
\a. \textbf{Premise:} Indeed, 58 percent of Columbia/HCA's beds lie empty, compared with 35 percent of nonprofit beds.
\b. \textbf{Hypothesis:} 58\% of Columbia/HCA's beds are empty, said the report. 
\c. \textbf{Label:} Entailment

\ex. \label{ex:hard2} \a. \textbf{Premise:} One he broke back to about the length of his forearm.
\b. \textbf{Hypothesis:} He snapped it until it was just a couple of inches long.
\c. \textbf{Label:} Contradiction

\ex. \label{ex:hard3} \a. \textbf{Premise:} The Kal nodded.
\b. \textbf{Hypothesis:} The Kal then shook its head side to side.
\c. \textbf{Label:} Contradiction

\ex. \label{ex:hard4} \a. \textbf{Premise:} Load time is divided into elemental and coverage related load time.
\b. \textbf{Hypothesis:} Load time is comprised of three parts.
\c. \textbf{Label:} Contradiction

\ex. \label{ex:hard5} \a. \textbf{Premise:} I thought working on Liddy's campaign would be better than working on Bob's.
\b. \textbf{Hypothesis:} I thought I would like working on Liddy's campaign the best.
\c. \textbf{Label:} Neutral

\ex. \label{ex:hard6} \a. \textbf{Premise:} Sure enough, there was the chest, a fine old piece, all studded with brass nails, and full to overflowing with every imaginable type of garment.
\b. \textbf{Hypothesis:} The chest wasn't big enough to completely contain all of the garments.
\c. \textbf{Label:} Entailment

\ex. \label{ex:hard7}
\a. \textbf{Premise:} True to his word to his faithful mare, Ca'daan left Whitebelly in Fena Dim and borrowed Gray Cloud from his uncle.
\b. \textbf{Hypothesis:} Ca'daan kept his word to Gray Cloud and borrowed Whitebelly from his uncle.
\c. \textbf{Label:} Contradiction

\ex. \label{ex:hard8} \a. \textbf{Premise:} Clearly, yes.
\b. \textbf{Hypothesis:} Obviously, the answer is yes.
\c. \textbf{Label:} Entailment

\noindent
Finally, examples \ref{ex:medium1} through \ref{ex:medium3} show some of the 8 cases that exactly half of our 100 instances got correct. Plausibly, such examples are the ones that lie close to a decision boundary that is relatively consistent across instances.

\ex. \label{ex:medium1} \a. \textbf{Premise:} He bent down to study the tiny little jeweled gears.
\b. \textbf{Hypothesis:} He bent down to examine the decorated gears.
\c. \textbf{Label:} Entailment

\ex. \label{ex:medium2} \a. \textbf{Premise:} Conversely, an increase in government saving adds to the supply of resources available for investment and may put downward pressure on interest rates.
\b. \textbf{Hypothesis:} Interest rates should increase to increase saving.
\c. \textbf{Label:} Contradiction

\ex. \label{ex:medium3} \a. \textbf{Premise:} More than 100 judges, lawyers and dignitaries were present for the gathering.
\b. \textbf{Hypothesis:} 152 judges and lawyers showed up
\c. \textbf{Label:} Neutral

\begin{figure*}
\centering
\resizebox{\textwidth}{!}{
\begin{tabular}{p{2cm}p{4.7cm}ccccc} \toprule
     Heuristic & Subcase & Minimum & Maximum & Mean & Std. dev. \\ \midrule
     Lexical & Untangling relative clauses & 0.94 & 1.00 & 0.98 & 0.01 \\
     overlap & \multicolumn{6}{l}{\textit{The athlete who the judges saw called the manager.} $\rightarrow$ \textit{The judges saw the athlete.}} \\
     \\
     & Sentences with PPs & 0.98 & 1.00 & 1.00 & 0.00 \\
     & \multicolumn{6}{l}{\textit{The tourists by the actor called the authors.}  $\rightarrow$ \textit{The tourists called the authors.}}\\ \\
     & Sentences with relative clauses & 0.97 & 1.00 & 0.99 & 0.01 \\
     & \multicolumn{6}{l}{\textit{The actors that danced encouraged the author.}  $\rightarrow$ \textit{The actors encouraged the author.}} \\ \\
     & Conjunctions & 0.72 & 0.92 & 0.83 & 0.05 \\
     & \multicolumn{6}{l}{\textit{The secretaries saw the scientists and the actors.} $\rightarrow$ \textit{The secretaries saw the actors.}}\\ \\
     & Passives & 0.99 & 1.00 & 1.00 & 0.00  \\ 
     & \multicolumn{6}{l}{\textit{The authors were supported by the tourists.} $\rightarrow$ \textit{The tourists supported the authors.}}\\ \\
     \midrule
     Subsequence & Conjunctions & 0.93 & 1.00 & 0.98 & 0.02 \\
     & \multicolumn{6}{l}{\textit{The actor and the professor shouted.} $\rightarrow$ \textit{The professor shouted.}}\\ \\
     & Adjectives & 1.00 & 1.00 & 1.00 & 0.00 \\
     & \multicolumn{6}{l}{\textit{Happy professors mentioned the lawyer.} $\rightarrow$ \textit{Professors mentioned the lawyer.}}\\ \\
     & Understood argument & 0.95 & 1.00 & 1.00 & 0.01 \\ 
     & \multicolumn{6}{l}{\textit{The author read the book.} $\rightarrow$ \textit{The author read.}} \\ \\
     & Relative clause on object & 0.98 & 1.00 & 0.99 & 0.01 \\ 
     & \multicolumn{6}{l}{\textit{The artists avoided the actors that performed.} $\rightarrow$ \textit{The artists avoided the actors.}}\\ \\
     & PP on object & 1.00 & 1.00 & 1.00 & 0.00 \\ 
     & \multicolumn{6}{l}{\textit{The authors called the judges near the doctor.} $\rightarrow$ \textit{The authors called the judges.}}\\ \\
     \midrule
     Constituent & Embedded under preposition & 0.81 & 1.00 & 0.96 & 0.02 \\ 
     & \multicolumn{6}{l}{\textit{Because the banker ran, the doctors saw the professors.} $\rightarrow$ \textit{The banker ran.}}\\ \\
     & Outside embedded clause & 1.00 & 1.00 & 1.00 & 0.00 \\
     & \multicolumn{6}{l}{\textit{Although the secretaries slept, the judges danced.} $\rightarrow$ \textit{The judges danced.}}\\ \\
     & Embedded under verb & 0.93 & 1.00 & 0.99 & 0.01 \\
     & \multicolumn{6}{l}{\textit{The president remembered that the actors performed.} $\rightarrow$ \textit{The actors performed.}}\\ \\
     & Conjunction & 1.00 & 1.00 & 1.00 & 0.00 \\ 
     & \multicolumn{6}{l}{\textit{The lawyer danced, and the judge supported the doctors.} $\rightarrow$ \textit{The lawyer danced.}}\\ \\
     & Adverbs & 1.00 & 1.00 & 1.00 & 0.00 \\
     & \multicolumn{6}{l}{\textit{Certainly the lawyers advised the manager.}  $\rightarrow$ \textit{The lawyers advised the manager.}}\\ \\
     \bottomrule
\end{tabular}
}
\caption{Results for the HANS subcases for which the heuristics make correct predictions (i.e., where the correct label is \textit{entailment}). All statistics are based on 100 runs.} \label{tab:fine_grained_entailed}
\end{figure*}

\begin{figure*}
\centering
\resizebox{\textwidth}{!}{
\begin{tabular}{p{2cm}p{4.7cm}ccccc} \toprule
     Heuristic & Subcase & Minimum & Maximum & Mean & Std. dev. \\ \midrule
     Lexical & Subject-object swap & 0.00 & 0.66 & 0.19 & 0.17 \\
     overlap & \multicolumn{6}{l}{\textit{The senators mentioned the artist.}  $\nrightarrow$ \textit{The artist mentioned the senators.}}\\ \\
      & Sentences with PPs & 0.04 & 0.76 & 0.41 & 0.18 \\
     & \multicolumn{6}{l}{\textit{The judge behind the manager saw the doctors.}  $\nrightarrow$ \textit{The doctors saw the manager.}}\\ \\
     & Sentences with relative clauses & 0.09 & 0.67 & 0.33 & 0.14 \\
     & \multicolumn{6}{l}{\textit{The actors called the banker who the tourists saw.} $\nrightarrow$ \textit{The banker called the tourists.}}\\ \\
     & Conjunctions & 0.12 & 0.72 & 0.45 & 0.15 \\
     & \multicolumn{6}{l}{\textit{The doctors saw the presidents and the tourists.} $\nrightarrow$ \textit{The presidents saw the tourists.}}\\ \\
     & Passives & 0.00 & 0.04 & 0.01 & 0.01 \\ 
     & \multicolumn{6}{l}{\textit{The senators were helped by the managers.} $\nrightarrow$ \textit{The senators helped the managers.}}\\ \\
     \midrule
     Subsequence & NP/S & 0.00 & 0.05 & 0.02 & 0.01 \\
     & \multicolumn{6}{l}{\textit{The managers heard the secretary resigned.} $\nrightarrow$ \textit{The managers heard the secretary.}}\\ \\ 
     & PP on subject & 0.00 & 0.35 & 0.12 & 0.07 \\
     & \multicolumn{6}{l}{\textit{The managers near the scientist shouted.} $\nrightarrow$ \textit{The scientist shouted.}}\\ \\
     & Relative clause on subject & 0.00 & 0.23 & 0.07 & 0.04 \\
     & \multicolumn{6}{l}{\textit{The secretary that admired the senator saw the actor.} $\nrightarrow$ \textit{The senator saw the actor.}}\\ \\
     & MV/RR & 0.00 & 0.02 & 0.00 & 0.00 & \\
     & \multicolumn{6}{l}{\textit{The senators paid in the office danced.} $\nrightarrow$ \textit{The senators paid in the office.}}\\ \\
     & NP/Z & 0.02 & 0.13 & 0.06 & 0.02 \\ 
     & \multicolumn{6}{l}{\textit{Before the actors presented the doctors arrived.} $\nrightarrow$ \textit{The actors presented the doctors.}}\\ \\
     \midrule
     Constituent & Embedded under preposition & 0.14 & 0.70 & 0.41 & 0.12 \\
     & \multicolumn{6}{l}{\textit{Unless the senators ran, the professors recommended the doctor.}  $\nrightarrow$ \textit{The senators ran.}}\\ \\
     & Outside embedded clause & 0.00 & 0.03 & 0.00 & 0.01 \\
     & \multicolumn{6}{l}{\textit{Unless the authors saw the students, the doctors resigned.} $\nrightarrow$ \textit{The doctors resigned.}}\\ \\
     & Embedded under verb & 0.02 & 0.42 & 0.17 & 0.08 \\
     & \multicolumn{6}{l}{\textit{The tourists said that the lawyer saw the banker.} $\nrightarrow$ \textit{The lawyer saw the banker.}}\\ \\
     & Disjunction & 0.00 & 0.03 & 0.00 & 0.01 \\
     & \multicolumn{6}{l}{\textit{The judges resigned, or the athletes saw the author.} $\nrightarrow$ \textit{The athletes saw the author.}}\\ \\
     & Adverbs & 0.00 & 0.17 & 0.06 & 0.04 \\ 
     & \multicolumn{6}{l}{\textit{Probably the artists saw the authors.}  $\nrightarrow$ \textit{The artists saw the authors.}}\\ \\
     \bottomrule
\end{tabular}
}
\caption{Results for the HANS subcases for which the heuristics make incorrect predictions (i.e., where the correct label is \textit{non-entailment}). All statistics are based on 100 runs.} \label{tab:fine_grained_nonentailed}
\end{figure*}

\begin{figure}[t]
    \centering
    \resizebox{\columnwidth}{!}{\begin{tabular}{p{4cm}cc} \toprule
        \textbf{Subject-object swap:}\newline The doctor visited the lawyer. $\nrightarrow$ The lawyer visited the doctor. & \raisebox{-0.9\totalheight}{\includegraphics[height=60px]{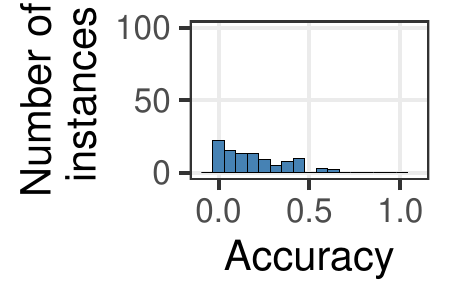}}   \\
        \textbf{Preposition:}\newline The tourist by the manager saw the artists. $\nrightarrow$ The artists saw the manager. & \raisebox{-0.9\totalheight}{\includegraphics[height=60px]{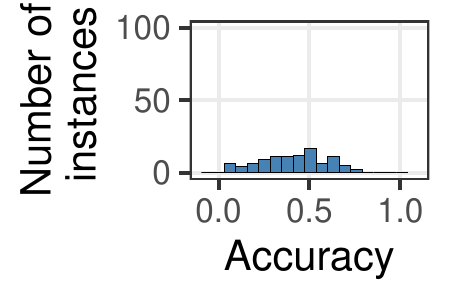}}  \\
        \textbf{Relative clause:}\newline The actors saw the author who the judge advised. $\nrightarrow$ The author saw the judge. & \raisebox{-0.9\totalheight}{\includegraphics[height=60px]{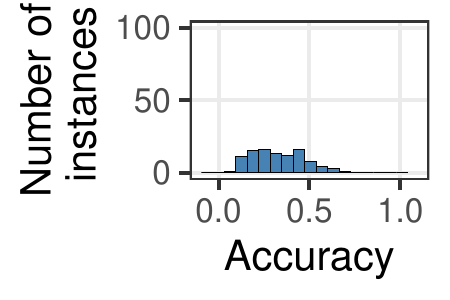}}\\
        \textbf{Passive:}\newline The student was stopped by the doctor. $\nrightarrow$ The student stopped the doctor. &  \raisebox{-0.9\totalheight}{\includegraphics[height=60px]{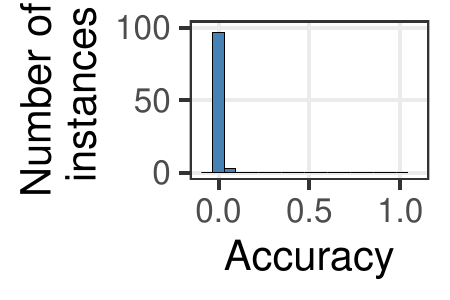}}\\
        \textbf{Conjunction:}\newline The doctors saw the athlete and the judge. $\nrightarrow$ The athlete saw the judge. & \raisebox{-0.9\totalheight}{\includegraphics[height=60px]{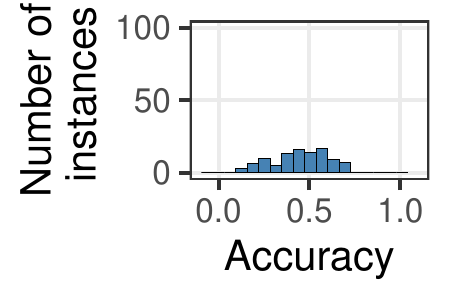}} \\ \bottomrule
    \end{tabular}}
    \caption{Accuracy distributions on the subcategories of the non-entailed lexical overlap examples of the HANS dataset (i.e., the examples that are inconsistent with the lexical overlap heuristic). For numerical results, and results for the other 25 subcategories of HANS, see Figures \ref{tab:fine_grained_entailed} and \ref{tab:fine_grained_nonentailed}.
    }
    \label{fig:subcategories}
\end{figure}

\subsection{Out-of-distribution generalization}

\begin{figure}
    \centering
    \includegraphics[width=\columnwidth]{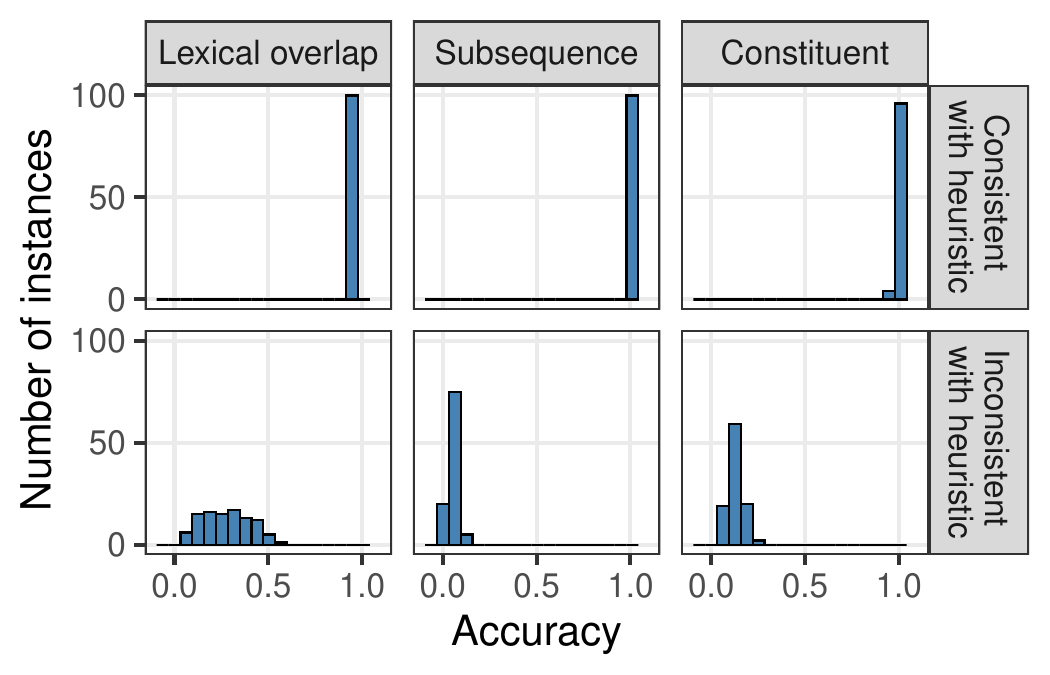}
    \caption{Out-of-distribution generalization: Performance on HANS, broken down into six categories of examples, based on the syntactic heuristic that each example targets and whether the example is consistent with the relevant heuristic (i.e., has a correct label of \textit{entailment}) or inconsistent with the heuristic (i.e., has a correct label of \textit{non-entailment}). The lexical overlap cases that are inconsistent with the heuristic (lower left plot) are highly variable across instances. For numerical results, see Figure \ref{tab:results_merged}.}
    \label{fig:hans_results}
\end{figure}

On HANS, performance was much more variable than on the MNLI development set. 
HANS consists of 6 main categories of examples, each of which can be further divided into 5 subcategories. 
Performance was reasonably consistent on five of these categories, 
but on the sixth category---lexical overlap examples that are inconsistent with the lexical overlap heuristic---performance varied dramatically, ranging from 5\% accuracy to 55\% accuracy (Figure \ref{fig:hans_results}). Since this is the most variable category, we focus on it for the rest of the analysis. 

The category of lexical overlap examples that are inconsistent with the lexical overlap heuristic encompasses examples for which the correct label is \textit{non-entailment} 
and for which all the words in the hypothesis also appear in the premise but not as a contiguous subsequence.
This category 
has five subcategories; examples and results for each subcategory 
are in Figure \ref{fig:subcategories}. Chance performance on HANS was 50\%; on all 
subcategories except for passives,
accuracies ranged from far below chance to modestly above chance. Models varied considerably even on categories that humans find simple \cite{mccoy2019right}. For example, 
accuracy on the subject-object swap examples, which can be handled with only rudimentary knowledge of syntax (in particular, the distinction between subjects and objects), ranged from 0\% to 66\%. Overall, although these models performed consistently on the in-distribution test set, they have nevertheless learned highly variable representations of syntax.

\setlength\tabcolsep{5pt}
\begin{figure*} [h]
\centering
\begin{tabular}{lccccccc}\toprule
&  & \multicolumn{3}{c}{HANS: Consistent with heuristic} & \multicolumn{3}{c}{HANS: Inconsistent with heuristic}\\
\cmidrule(lr){3-5} \cmidrule(lr){6-8}

 & MNLI & Lexical & Subseq. & Const. & Lexical & Subseq. & Const.   \\ \midrule
Minimum & 0.84 & 0.93 & 0.98 & 0.96 & 0.05 & 0.01 & 0.03 \\
Maximum & 0.85 & 0.98 & 1.00 & 1.00 & 0.55 & 0.14 & 0.24 \\
Mean & 0.84 & 0.96 & 0.99 & 0.99 & 0.28 & 0.05 & 0.13 \\
Standard deviation & 0.00 & 0.01 & 0.00 & 0.01 & 0.12 & 0.02 & 0.04 \\
\bottomrule
\end{tabular}
\caption{Results for models trained on MNLI. The MNLI column reports accuracy on the MNLI \texttt{matched} development set, where there are three possible labels (\textit{entailment}, \textit{contradiction}, and \textit{neutral}). The remaining columns are subsets of the HANS dataset, with \textit{neutral} and \textit{contradiction} merged into a single label, \textit{non-entailment}, such that there are only two possible labels: \textit{entailment} and \textit{non-entailment}. The examples that are consistent with the heuristics are those that have a correct label of \textit{entailment}, while the examples that are inconsistent with the heuristics are those with a correct label of \textit{non-entailment}. All statistics are based on 100 runs. } \label{tab:results_merged}
\end{figure*}
\setlength\tabcolsep{6pt}

\section{Discussion}

We have found that models that differ only in their initial weights and the order of training examples can vary substantially in out-of-distribution linguistic generalization. We found this variation even with the vast majority of initial weights held constant (i.e., all the weights in the BERT component of the model).
We conjecture that models might be even more variable
if the pre-training of BERT
were also redone across instances.
These results underscore the importance of evaluating models on multiple restarts, as conclusions drawn from a single instance of a model might not hold across instances. Further, these results highlight the importance of evaluating out-of-distribution generalization; since all of our instances displayed similar in-distribution generalization, only their  
out-of-distribution generalization  
illuminates the substantial differences in what they have learned.

In stark contrast to the models we have looked at---which generalized in highly variable ways despite being trained on the same set of examples---humans tend to converge to similar linguistic generalizations despite major differences in the linguistic input that they encounter as children \cite{chomsky1965,chomsky1980}. This suggests that reducing the generalization variability of NLP models 
may help bring them closer to human performance in one major area where they still dramatically lag behind humans, namely in out-of-distribution generalization.

How could the out-of-distribution generalization of 
models be made more consistent? The variability that we have observed likely reflects the presence of many local minima in the loss surface, all of which are equally attractive to our models. 
This makes the model's choice of a minimum essentially arbitrary and easily affected by the initial weights and the order of training examples. 
To reduce this variability, then, one approach would be to use models with stronger inductive biases, which can help distinguish between the many 
local minima. An alternate approach would be to use training sets that better represent a large set of linguistic phenomena, to decrease the probability of there being local minima that ignore certain phenomena.

\section*{Acknowledgments}

We are grateful to Emily Pitler, Dipanjan Das, and
the members of the Johns Hopkins Computation
and Psycholinguistics lab group for helpful comments. Any errors are our own.

This project is based upon work supported
by the National Science Foundation Graduate
Research Fellowship Program under Grant No.
1746891 and by a gift to TL from Google, and it was
conducted using computational resources from the
Maryland Advanced Research Computing Center
(MARCC). Any opinions, findings, and conclusions or recommendations expressed in this material are those of the authors and do not necessarily
reflect the views of the National Science Foundation, Google, or MARCC.

\bibliographystyle{acl_natbib}
\bibliography{feather}

\end{document}